\def\year{2019}\relax
\documentclass[letterpaper]{article} 
\usepackage{aaai19}  
\usepackage{times}  
\usepackage{helvet}  
\usepackage{courier}  
\usepackage{url}  
\usepackage{graphicx}  
\usepackage{algorithm, algpseudocode}
\begin{document}
%
\title{Deep Multimodal Learning: An Effective Method for Video Classification}
\author {Tianqi Zhao\\
Institute for Interdisciplinary Information Sciences\\
Tsinghua University\\
zhaotq16@mails.tsinghua.edu.cn}
\maketitle
\begin{abstract}
Videos have become ubiquitous on the Internet. And video analysis can provide lots of information for detecting and recognizing objects as well as help people understand human actions and interactions with the real world. However, facing data as huge as TB level, effective methods should be applied. Recurrent neural network (RNN) architecture has wildly been used on many sequential learning problems such as Language Model, Time-Series Analysis, etc. In this paper, we propose some variations of RNN such as stacked bidirectional LSTM/GRU network with attention mechanism to categorize large-scale video data. We also explore different multimodal fusion methods. Our model combines both visual and audio information on both video and frame level and received great result. Ensemble methods are also applied. Because of its multimodal characteristics, we decide to call this method Deep Multimodal Learning(DML). Our DML-based model was trained on Google Cloud and our own server and was tested in a well-known video classification competition on Kaggle held by Google.
\end{abstract}

\section{Introduction} 
Today people are able to watch a tremendous amount of videos, both on television and the Internet. The increasing amount of videos make a viewer difficult to find his or her favorite video immediately. One method that viewers use to narrow their choices is to look for videos within specific categories. In fact, almost 300 hours of video are uploaded to YouTube\footnote{\url{www.youtube.com}} , one of the biggest video websites in the world, every minute. Due to the huge amount of videos to classify, it is impossible to categorize them manually. The request for categorizing them has encouraged the development of research on large-scale video classification automatically.

Video classification can be briefly described as that given the pre-extracted both audio and visual features of videos on video-level as well as frame-level, we want to assign the right video-level label(s) for them. This task is much more difficult than picture classification. For the first reason, video consists of a sequence of pictures (also called frames) and audio data, which makes the number of features remarkably large. As the number of features increases, model will be hard to train. The second reason is that data redundancy widely exists in video data. Maybe it is because that successive frames have little differences, and similar frames provide no help to classification. The last reason is, each picture always has one unique label for training. However, several labels will be attached to one video, since a video can include lots of contents. 

Recent studies have developed new models and techniques for the video classification problem, and how to efficiently deal with the {\em large-scale} video categorization problem has begun to interest researchers. Though Convolutional Neural Networks (CNNs) have been extensively applied for image recognition problems \cite{yue2015beyond}, but it scarcely captures the characterization of time sequence. In this paper, we adopt a Recurrent Neural Network (RNN). RNNs are able to make use of sequential information, suitable for this problem. And we use some techniques to strengthen the ability of digging out relationship between frames.
\subsection{Our Contribution}
In this paper, we introduce a method called Deep Multimodal Learning(DML) for the video classification problem that is compatible with large-scale tasks. The whole model is based on RNN. We test different variations of it such as stacked bi-LSTM and stacked bi-GRU as well as attention mechanism. We also explore three different multimodal fusion techniques. We compare the performance of different techniques by using video-level and frame-level features. Our model is implemented and tested in a video classification competition, \emph{Google Cloud \& YouTube-8M Video Understanding Challenge}\footnote{\url{https://www.kaggle.com/c/youtube8m}} on Kaggle. Our best rank is 90 over 656 participating teams at the deadline of the competition. The score we obtain is \emph{0.80583} while the winner team is \emph{0.84967}. We also propose some methods for improvement (using Ensemble Learning) and come up with some methods to improve our model in the future work.
\section{Related Work}
Image datasets are very important in many fields in computer vision, such as MNIST\cite{lecun1998gradient} and CIFAR-10\cite{krizhevsky2009cifar}. There are also some bigger datasets, like ImageNet\cite{deng2009imagenet} and Microsoft COCO\cite{lin2014microsoft}. In the video understanding domain, we can see a similar progress. Starting from KTH\cite{laptev2005space}, Hollywood 2\cite{laptev2008learning}, with a few thousand video clips, to the medium size dataset like UCF101\cite{soomro2012ucf101}, however, none of them is as huge as YouTube-8M that we use. Therefore, YouTube-8M shows its importance because it serves as a benchmark in the video understanding area.

Multimodal is an emerging area which focuses on using various modalities to improve the performance of the model. More specifically, it tends to solve these five problems: Representation, Translation, Alignment, Fusion and Co-learning. Representation means learning how to represent and summarize multimodal data that exploits the complementarity and redundancy of multiple modalities. Translation studies how to translate data from one modality to another. Alignment tries to identify the direct relations between elements from over two modalities. Fusion is a challenge joining information from two or more modalities to perform a prediction. Co-learning transfers knowledge between modalities, their representation, and their predictive models\cite{baltruvsaitis2018multimodal}. What we concern in our work is Fusion, cause the whole task is about prediction with two modalities: visual and audio. It includes two types of methods: Joint and Coordinated. We mainly cares about Joint method, which consists of three types of methods, including neural networks, graphical models and sequential. We care about the first one because our model adopts end-to-end neural networks.  

Since year 2015, a mechanism called attention is widely used in several fields in natural language processing area. For example, attention is applied in machine translation in \cite{bahdanau2014neural}, which significantly improve the BLEU score in machine translation due to the fact that when translating, the decoder knows the exact place to pay attention to. Attention can also be used when reasoning about entailment\cite{rocktaschel2015reasoning}, which helps to improve the accuracy of classification. In the computer vision area, attention can also be used in the task of image caption\cite{xu2015show}, by focusing on the crucial part of an image, the caption is able to describe a picture more accurately.

Recently, some fancy methods are applied to large video classification with visual and audio features and obtain good performance.  \cite{li2017temporal} use Temporal Resnet Blocks(TRB), each TRB consists of two temporal convolutional layers (followed by batch norm and activation), and this structure is followed by a 7-layer Bi-LSTM/Bi-GRU with short-cut connections. \cite{wang2017truly} introduces a new text modality by crawling the title and keywords of youtube videos to enrich the dataset. \cite{miech2017learnable} adopts Vector of Locally aggregated Descriptors(VLAD)\cite{jegou2010aggregating} or Fisher Vectors(FV)\cite{perronnin2007fisher} as a way for pooling features.
\cite{wang2017monkeytyping} proposes a network structure called chaining which separates training data into two parts and train the model in two stages. \cite{zou2017youtube} uses variations of Long Short Term Memory(LSTM) model with some tricks and ensemble techniques to get good results. 
\section{Our Method}
We tried some variations of stacked RNN and explores several ways of multimodal fusion, besides, we adopt some techniques of Natural Language Processing such as attention mechanism. Because our model combines both visual and audio information on both video and frame level, we decide to call this method Deep Multimodal Learning(DML).
\subsection{Long Short-Term Memory}
A recurrent neural network (RNN) is a type of neural network architecture particularly suited for modeling sequential phenomena\cite{kim2016character}. RNNs can use their internal memory to process arbitrary sequences of inputs. $x_t$ is the input and $s_t$ is the hidden state at time step $t$. $s_t$ is the memory of the network, which is calculated based on the previous hidden state and the input at the current step, namely 
\[
    s_t = f(Ux_t + Ws_{t-1})
\]
where the function $f$ usually is nonlinear and the first hidden state $s_0$ is typically initialized to all zeros. $U$ and $W$ are the parameters needed to be trained. Unlike a traditional deep neural network, which uses different parameters at each layer, a RNN shares the same parameters across all steps. See Figure 1\cite{lecun2015deep}:

\begin{figure}[h!]
	\centering
	\includegraphics[width=0.4\textwidth]{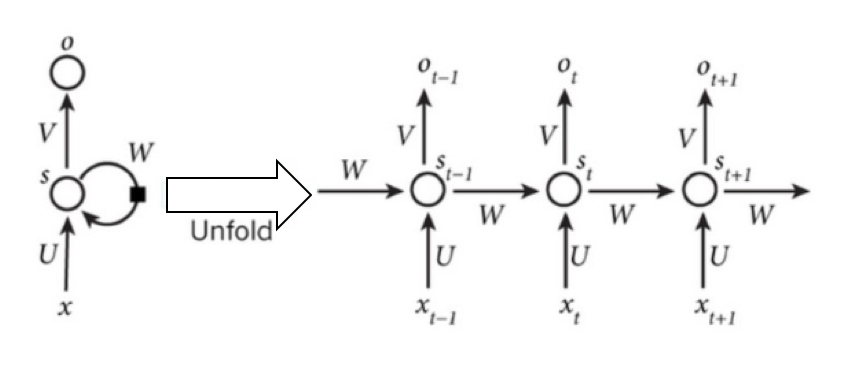}
	\caption{An unfolded \emph{Recurrent Neural Network}}
	\label{fig:lstm}
\end{figure}

We can see from the picture above that:

$x_t$ is the input at time step t. $s_t$ is the hidden state at time step t. The function f usually is a nonlinearity such as tanh or ReLU. First hidden state is typically initialized to all zeroes. $o_t$ is the output at step t. If we wanted to predict the next word in a sentence it would be a vector of probabilities across our vocabulary. And we use this formula:
\[
	o_t =\mathrm{softmax}(Vs_t).
\]

Long Short-Term Memory (LSTM) \cite{hochreiter1997long} networks are a special kind of RNN, capable of learning long-term dependencies. An LSTM network is well-suited to learn from experience to classify, process and predict time series when there are time lags of unknown size and bound between important events\cite{lyu2014revisit}.

The structure of a traditional LSTM cell (block), shown in Figure 2, is straightforward. $x_t$ is the input vector and $h_t$ is the output vector. LSTM blocks contain three or four "gates" that they use to control the flow of information into or out of their memory. These gates are implemented using the logistic function to compute a value between 0 and 1. Multiplication is applied with this value to partially allow or deny information to flow into or out of the memory.

A LSTM unit can be described as below:
\begin{itemize}
	\item Variables
	\begin{itemize}
		\item $x_t$: input vector
		\item $h_t$: hidden state vector
		\item $c_t$: cell state vector
		\item $f_t$, $i_t$, $o_t$: gate vectors
		\begin{itemize}
			\item $f_t$: Forget gate vector. Weight of remembering old information.
			\item $i_t$: Input gate vector. Weight of acquiring new information.
			\item $o_t$: Output gate vector. Output candidate.
		\end{itemize}
	\end{itemize}
	\item Parameters: $W$, $b$
	\item Activation functions: $\sigma_g$, $\sigma_c$, $\sigma_h$
	\item Data transfer formulas:
	\\
	\begin{eqnarray}
			&i_t&= \sigma_g (W_{xi}x_t + W_{hi}h_{t-1} + W_{ci}c_{t-1} + b_i) \nonumber
			\\
			&f_t&= \sigma_g (W_{xf}x_t + W_{hi}h_{t-1} + W_{cf}c_{t-1} + b_f) \nonumber
			\\
			&c_t&= f_t\times c_{t-1} + i_t \times \sigma_c(W_{xc}x_{t} + W_{hc}h_{t-1} + b_c) \nonumber
			\\
			&o_t&= \sigma_g (W_{xo}x_t + W_{ho}h_{t-1} + W_{co}c_{t} + b_o) \nonumber
			\\
			&h_t&= o_t \times \sigma_h(c_t) \nonumber
	\end{eqnarray}
\end{itemize}
Since RNN is trained by back propagation through time(BPTT), and therefore unfolded into feed forward net with multiple layers. When gradient is passed back through many time steps, it tends to explode or vanish, however, LSTM is able to alleviate this problem, thanks to its unique cell structure. This is the reason why we choose LSTM instead of vanilla RNN.
\subsection{Bi-LSTM}
Since we use the frame-level features to train our model, traditional LSTM is able to capture the current frame's dependencies on previous frames. This implies that the current frame is relevant to frames ahead of it. We intuitively think that the current frame is also largely dependent on the following frames. Therefore, we adopt a bidirectional LSTM instead of traditional LSTM. Bidirectional LSTM networks can significantly improve the performance of classification and recognition, because after calculating the forward hidden state, it calculates backward hidden states, which extracts the internal relationship between frames of both directions.
\begin{figure}[h!]
	\centering
	\includegraphics[width=0.5\textwidth]{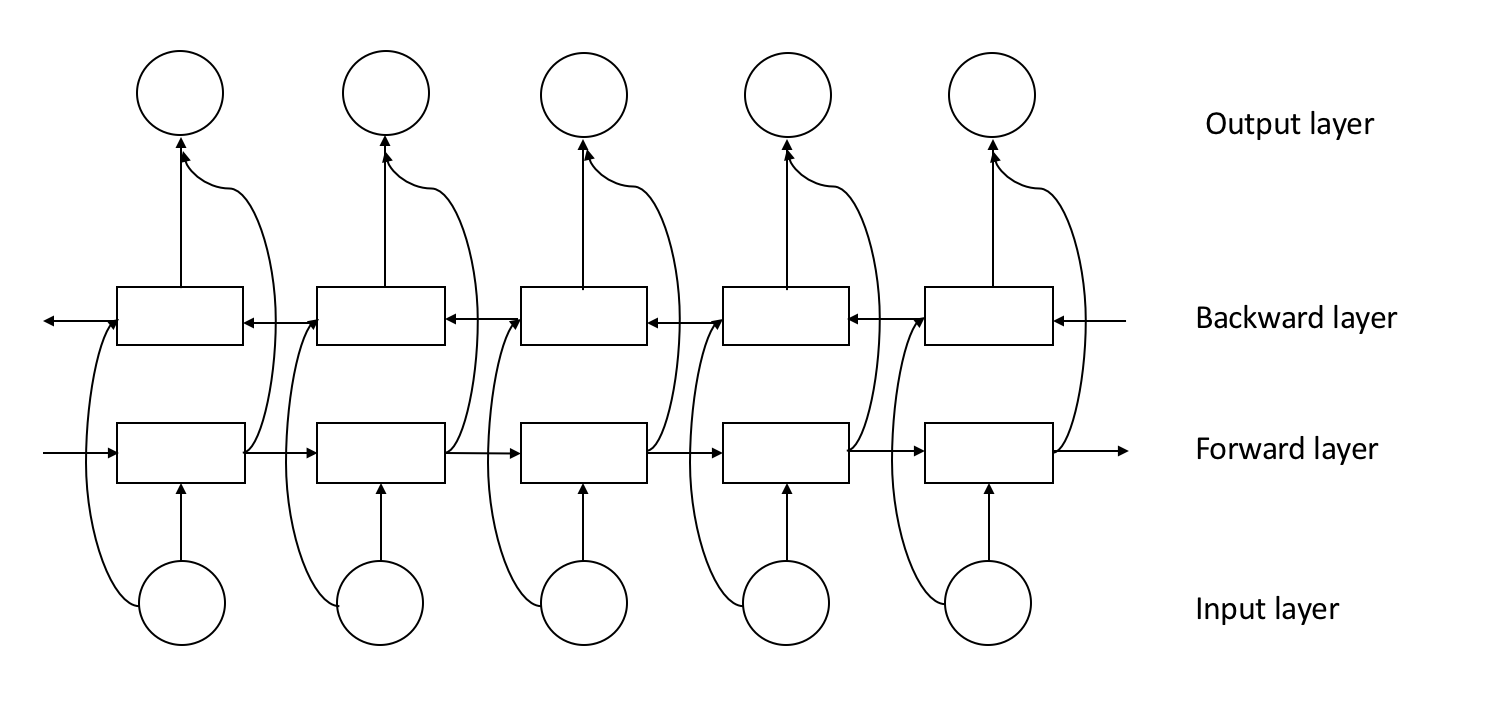}
	\caption{Bi-LSTM architecture using both visual and audio features}
	\label{fig:arch}
\end{figure}
We use a two-layer stacked bi-lstm model, for each direction, the model uses two layers to improve the representation power of feature extraction. Too many layers will increase the model complexity and therefore slow the training speed. Therefore a two-layer bi-lstm model is suited for this task.

\subsection{Bi-GRU}
The biggest difference between Bi-GRU model and Bi-LSTM model is the unit. Bi-GRU replaces a LSTM unit with a Gated Recurrent Unit (GRU). The mechanism is as follows:
	\begin{eqnarray}
			&r_t&= \sigma (W_{r}[h_{t-1}, x_t]) \nonumber
			\\
			&z_t&= \sigma (W_{z}[h_{t-1},x_t]) \nonumber
			\\
			&\hat h_t&= tanh(W_{\hat h}[r_t * h_{t-1}, x_t]) \nonumber
			\\
			&h_t&= (1-z_t) * h_{t-1} + z_t * \hat h_t \nonumber
			\\
			&y_t&= \sigma(W_o h_t) \nonumber
	\end{eqnarray}
where the hidden state $h_t$ takes the role of memory, reset gate $r_t$ decides how much of the previous memory should be remembered, and update gate $z_t$ controls how much of the previous memory content is to be forgotten and how much of the new memory is to be added. The notation * represents element-wise production, and W is the parameter need to be trained.

\subsection{Multimodal Fusion}
A video contains both audio and visual information and they are intimately correlated in time structure and content helping us understand the video better, which suggests we need to try a way to make the best use of them. 

To speak specifically, find a best way to fuse different modalities. In this paper, we tried three different ways to combine them. 
Firstly, we try to concatenate them directly as a whole feature vector:
\[
x_m = [x_1;x_2;...;x_n]
\]
Secondly, we try to combine the unimodal signals into the same representation space(shared space method):
\[
x_m = f(x_1,...,x_n)
\]
Thirdly, we try to project them into the spaces of same dimensions(projection method), so that the different modalities of feature vectors are close under the distance of $L_2$ norm:
\[
f(x_1) \sim g(x_2)
\]
The different architectures are shown as Figure 3. 
\begin{figure}[h!]
	\centering
	\includegraphics[width=0.5\textwidth]{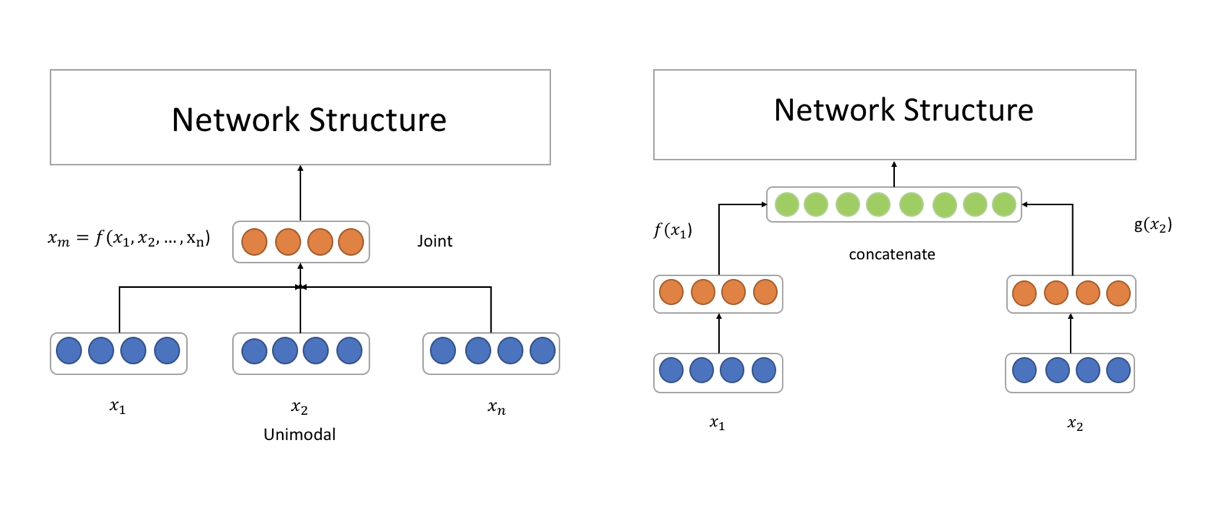}
	\caption{Multimodal fusion}
	\label{fig:arch}
\end{figure}

\subsection{Attention}
We adopt the method of \cite{yang2016hierarchical}, which designs a special attention mechanism for text classification.
\begin{figure}[h!]
	\centering
	\includegraphics[width=0.4\textwidth]{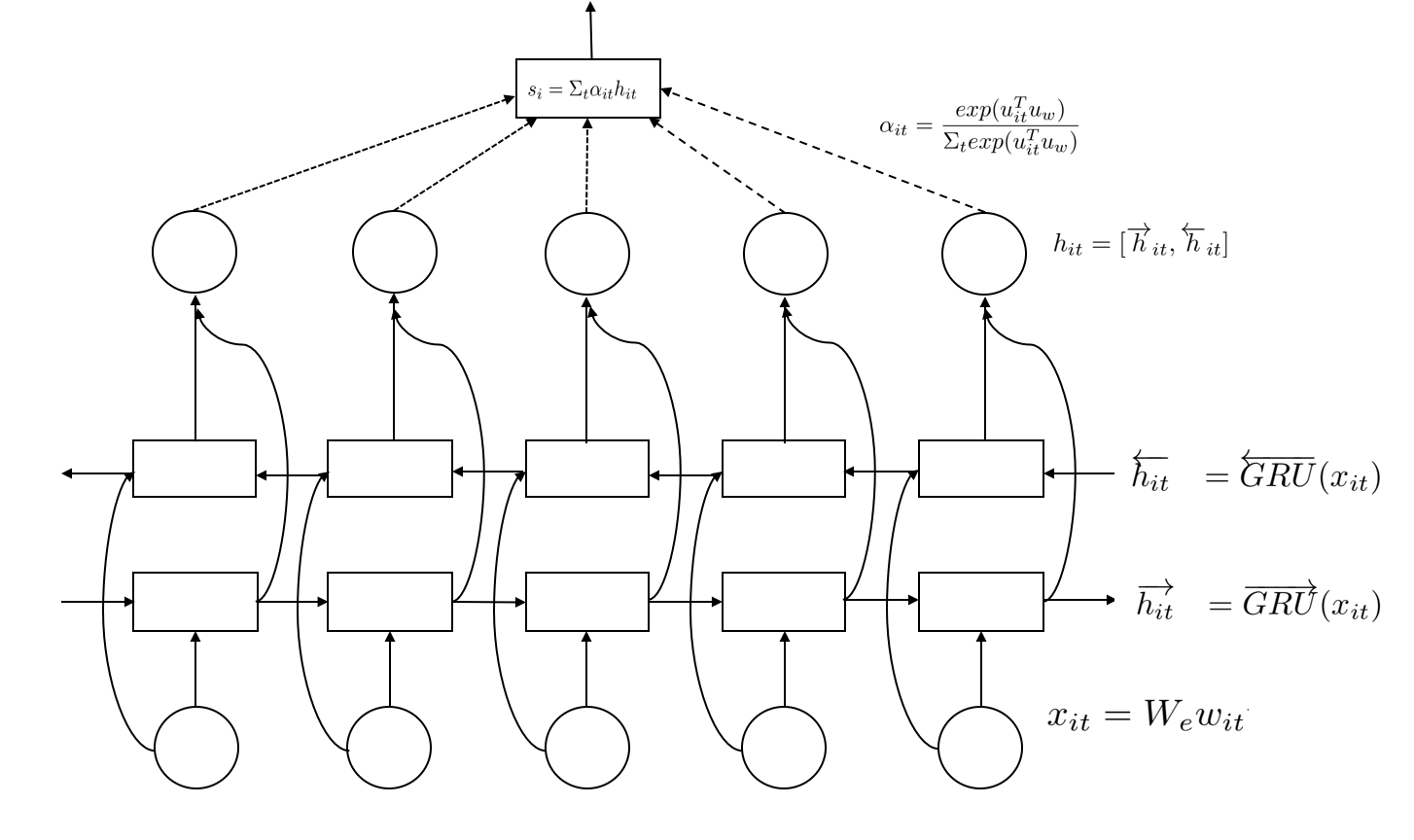}
	\caption{Bi-LSTM model with attention machenism}
	\label{fig:arch}
\end{figure}

As Figure 4, given a frame sequence $s_i$, we represent frames as $[w_i1, w_i2, w_i3,...]$, and we use a embedding matrix to encode the frames.
\[
x_{it} = W_e w_{it} t\in[1,T]
\]
where $W_e$ is the embedding matrix.

We use bi-GRU to encode the hidden state of every frame:
	\begin{eqnarray}
			&\overrightarrow {h_{it}}&= \overrightarrow{GRU} (x_{it}) \nonumber
			\\
			&\overleftarrow{h_{it}}&= \overleftarrow{GRU} (x_{it}) \nonumber
	\end{eqnarray}
And finally the we get the final hidden state by concatenating these two hidden states:
\[
h_{it} = [\overrightarrow h_{it}, \overleftarrow h_{it}]
\]
After that, we use a multilayer perceptron(MLP) to get the representation of $h_{it}$:
\[
u_{it} = tanh(W_{w} h_{it} + b_w)
\]
and use softmax function for normalization:
\[
\alpha_{it} = \frac{exp(u_{it}^T u_w)}{\Sigma_t exp(u_{it}^T u_w)}
\]
Finally we calculate the weighted average of hidden states, and get the representation of a sequence:
\[
s_i = \Sigma_t \alpha_{it} h_{it}
\]
This attention mechanism can help the model attend on the certain parts of the frames, and get more information from a certain frame sequence.

\subsection{Optimization}
We use Adam optimizer\cite{kingma2014adam} to learn the parameters of our model. Adam(Adaptive Moment Estimation) is essentially a RMSprop with Momentum term\cite{duchi2011adaptive}. It is suitable for big data set and high dimension space.
The formulas are as follows:
\begin{eqnarray}
&m_t&=\mu \times m_{t-1}+(1-\mu) \times g_t\\
&n_t&=v \times n_{t-1}+(1-v)\times g_t^2\\
&m_t&=\frac{m_t}{1-\mu^t}\\
&n_t&=\frac{n_t}{1-v^t}\\
&\Delta\theta_t&= -\frac{\hat m_t}{\sqrt{\hat n_t ̂+ \epsilon}}
\end{eqnarray}
where $\mu$ represents the exponential decay rate for the first moment estimates and v represents the exponential decay rate for the second-moment estimates. $\epsilon$ is a pretty small number to prevent zero division. And it can be easily seen that for each time step the $m_t$ and $n_t$ is updated.

According to what the paper\cite{kingma2014adam} suggests, the Adam algorithm is as follows:
\begin{algorithm}[htb]
	\caption {Adam is our chosen algorithm for stochastic optimization. $g^{2}_{t}$ indicates the elementwise square of $g_t$. We set $\alpha = 0.001$, $\mu = 0.9$, $v = 0.999$,$\epsilon = 10^{−8}$, which is the default setting that the paper suggests. All operations on vectors are element-wise. And $\mu^t$ and $v^t$ represent $\mu$ and $v$ to the power t.}
	\label{alg:Adam}
	\begin{algorithmic}[1]
		\Require $\alpha$:Stepsize
		\Require $\mu, v\in[0,1)$: Exponential decay rates for the moment estimates
		\Require $f(\theta)$:Stochastic objective function with parameters $\theta$
		\\
		$\theta_0$:Initial parameter vector
		\\
		$m_0\gets$ 0 (Initialize $1^{st}$ moment vector)
		\\
		$v_0\gets$ 0 (Initialize $2^{nd}$ moment vector)
		\\
		$l\gets$ 0 (Initialize time step)
		\While {$\theta_t$ not converged}
		\State $t\gets t + 1$
		\State $g_t\gets\nabla_\theta f_t(\theta_{t-1})$(Get gradients w.r.t. stochastic objective at timestep t)
		\State $m_t\gets\mu \cdot m_{t-1} + (1-\mu) \cdot g_t$(Update biased ﬁrst moment estimate)
		\State $n_t\gets v \cdot n_{t-1} + (1-v) \cdot g_t^2$(Update biased second raw moment estimate)
		\State $\hat m_t \gets \frac{m_t}{1-\mu^t}$(Compute bias-corrected first moment estimate)
		\State $\hat n_t \gets \frac{n_t}{1-v^t}$(Compute bias-corrected second moment estimate)
		\State $\theta_t \gets \theta_{t-1}-\alpha\cdot\frac{\hat m_t}{(\sqrt{\hat n_t} + \epsilon)}$(Update parameters)
		\EndWhile
		\\
		\Return $\theta_t$(Resulting parameters)
	\end{algorithmic}
\end{algorithm}

Adam combines the benefits of both AdaGrad and RMSProp so that it can handle sparse gradients on noisy problems. Besides, the property that it is relatively easy to configure(the default configuration is powerful enough to handle most problems) is another reason why we choose it.
\subsection{Ensemble Learning}
In the end of our experiment, we will introduce ensemble learning to slightly improve the performance of our model.An ensemble contains several learners, we call them base learners according to the convention. By combining several weak learners which are slightly better than random guess, the strong learners that they get can make very accurate predictions. Base learners are usually learned from training data by a base learning algorithm which can be decision tree, logistic regression, neural networks or other kinds of machine learning algorithms. There are many ways to combine the base learners, such as majority voting for classification or weighted averaging for regression.

We design the following algorithm for ensemble learning, as follows:
\begin{algorithm}[htb]
	\caption {Ensemble of weak learners can provide a strong learner.}
	\label{alg:Ensemble}
	\begin{algorithmic}[1]
		\Require $\alpha_i \in [0,1)$, $\alpha_i = \frac{GAP_i}{\Sigma GAP_i}, \Sigma \alpha_i = 1$, $i$ is model number.
		\For {each video}
			\For {each class}
				\For {each model}
					\State $S_i \leftarrow$ GAP in model i
				\EndFor
				\State New score for the class $\leftarrow \Sigma S_i \times \alpha_i$
			\EndFor
			\State Sort class label according to the score value
			\State Keep top k results
		\EndFor
		\\
		\Return Top k labels for all videos
	\end{algorithmic}
\end{algorithm}

\section{Experiments}
We evaluate our model on the benchmark provided by the competition \emph{Google Cloud \& YouTube-8M Video Understanding Challenge}. In this competition, participants are challenged to develop classification algorithms which accurately assign video-level labels using the new and improved Youtube-8M\footnote{\url{https://research.googleblog.com/2016/09/announcing-youtube-8m-large-and-diverse.html}} dataset. All participants could sign up for a \emph{Google Cloud} free trial account which  includes \$300 in credits. The funding constraint is an obstacle of our experiments and we must train and test our model accurately.
\subsection{Dataset}
Today, one of the greatest obstacles to rapid improvements in video understanding research has been the lack of large-scale, labeled datasets open to the public. Google’s recent release of the \emph{YouTube-8M} (YT-8M) dataset represents a significant step in this direction. Their availability has significantly accelerated research in areas such as representation learning and video modeling architectures. We will apply our methods on this dataset.

The dataset was created from over 8 million YouTube videos (500,000 hours of video, see Figure 4) and includes video labels from a vocabulary of 4716 classes (3.4 labels/video on average), see Figure 5. According to the white paper\cite{abu2016youtube}, YouTube-8M use Knowledge Graph entities to succinctly describe the main themes of a video. For example, a video shows that a person is biking on dirt roads and cliffs would have a central topic of Mountain Biking, not Dirt, Road, Person, Sky, and so on. Therefore, the aim of the dataset is not only to understand what is present in each frame of the video, but also to identify the few key topics that best describe what the video is about. This would produce thousands of labels on each video but without answering what the video is really about. It also comes with pre-extracted audio \& visual features from every second of video (3.2B feature vectors in total). This represents a significant increase in scale and diversity compared to existing video datasets.

\begin{figure}[h!]
	\centering
	\includegraphics[width=0.4\textwidth]{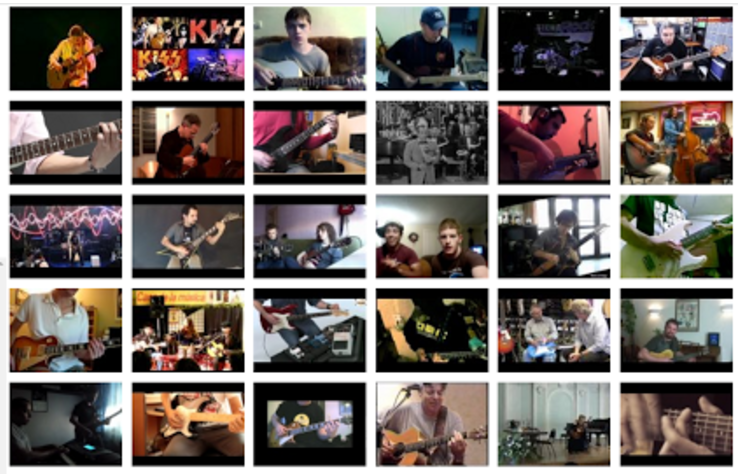}
	\caption{Some videos of \emph{YouTube-8M}}
	\label{fig:yt8m}
\end{figure}

\subsubsection{Video-Level Data}
The video-level data is 31GB, each video has four kinds of properties: video-id, labels, mean\_rgb, mean\_audio.
\begin{itemize}
	\item "video\_id": unique id for the video, in train set it is a Youtube video id, and in test/validation they are anonymized.
	\item "labels": list of labels of that video.
	\item "mean\_rgb": float array of length 1024.
	\item "mean\_audio": float array of length 128.
\end{itemize}
\subsubsection{Frame-Level Data}
The frame-level data is 1.71TB, each video has four kinds of properties: video-id, labels, rgb for each frame, audio for each frame.
\begin{itemize}
	\item "video\_id": unique id for the video, in train set it is a Youtube video id, and in test/validation they are anonymized.
	\item "labels": list of labels of that video.
	\item "rgb": float array of length 1024.
	\item "audio": float array of length 128.
\end{itemize}

\subsection{Environment}
We choose the \textbf{Google Cloud} platform to conduct our experiments due to the huge amount of TB-level data. Google Cloud Platform is a cloud platform that consists of a set of physical assets, such as computers and hard disk drives, and virtual resources, such as virtual machines (VMs), that are contained in Google's data centers around the globe. Like many other cloud platforms, it provides lots of high-quality service, such as Compute Engine, Storage and Databases, Machine Learning APIs. This distribution of resources provides several advantages, including redundancy in case of failure and reduced latency by locating resources closer to clients. Therefore, the platform is really helpful during our training. And the framework we choose is the well-known deep learning framework, naming \textbf{Tensorflow}, created by Google. 

\subsection{Approach}
We first execute a logistic regression (LR) model with video-level features and Long Short-Term Memory(LSTM) with frame-level features as baselines, which are provided by the starter code. Then we tried to make our model more complex to improve the representation power by using stacked Bi-LSTM and stacked Bi-GRU. After that, we explore three different multimodal fusion techniques to test which is the best. Finally, we introduce a natural language processing technique called attention mechanism to let the model pay more attention on crucial frames. 

The learning algorithm is Adam \cite{kingma2014adam}, which is an algorithm for first-order gradient-based optimization of stochastic objective functions, based on adaptive estimates of lower-order moments.

\subsection{Evaluation}
Our submissions are evaluated according to the \emph{Global Average Precision} (GAP) at $k$, where $k=20$. For each video, we submit a list of predicted labels and corresponding confidence scores. The evaluation takes the predicted labels that have the highest $k$ confidence scores for each video, then treats each prediction and the confidence score as an individual data point in a long list of global predictions, to compute the Average Precision across all of the predictions and all the videos. 

If a submission has $N$ predictions (label/confidence pairs) sorted by its confidence score, then the Global Average Precision is computed as:
\[
	GAP = \sum_{i=1}^{N}p(i)\Delta r(i)
\]
where $N$ is the number of final predictions (if there are 20 predictions for each video, then $N = 20 \times number \ of \ videos$), besides, $p(i)$ is the precision, and $r(i)$ is the recall.  

\subsection{Result}
Firstly we tried the Logistic Regression method and use this result as a bench mark, we used one GPU on the \emph{Google Cloud} and spend totally 3 hours on LR model. We know that some team try to use LSTM method and the GAP value has been improved a lot compared to the benchmark. After that, we tried to improve our model by introducing Bi-LSTM network and use some techniques to adjust the parameters, we use 4 GPUs and after over 155 hours we reached our best rank, which is 90 over 656 participating teams in the competition. The score we obtain is \emph{0.80583} while the winner team is \emph{0.84967}. After the competition, we also tried Bi-GRU and several multimodal fusion techniques, as well as attention mechanism and ensemble method. The learning rate for Adam is 0.01 and will have a decay of $95 \%$ after 4000000 steps, when the loss nearly converges, the steps becomes 40000 for each decay. For the mapping weight matrix in modality fusion experiments, we adept normal distribution with 0 mean and 0.01 dev. While the weight matrix used in attention model tries glorot normal initializer\cite{glorot2010understanding} which is proved to have better performance. See the results below: 
\begin{center}
	The GAP of Models
	\\
	\begin{tabular}{|c|c|c|c|c|}\hline
		\textsc{model}& GAP Score\\ \hline
		\textbf{Bi-GRU, shared-space, frame-level} & 0.01106 \\ \hline
		\textbf{Bi-GRU, projection, frame-level} & 0.01684 \\ \hline
		LR, video-level  & 0.70727 \\ \hline
		LSTM, frame-level  & 0.79903 \\ \hline
		\textbf{Bi-LSTM, frame-level}  & 0.80597 \\ \hline
		\textbf{Attention, frame-level} & 0.81415 \\ \hline
		\textbf{Ensemble, LR+LSTM}  & 0.81891 \\ \hline
		\textbf{Bi-GRU, concat, frame-level} & 0.84022 \\ \hline	
		{State-of-art, frame-level} & 0.84967 \\ \hline	
	\end{tabular}
\end{center}

\section{Discussion}
According to the experiments we conduct, we can draw a conclusion that the GAP value is improving with the model becoming more complicated.
\subsection{From LR to LSTM}
Logistic Regression is the most easy-to-think way to perform classification tasks. Thus it serves as a baseline. We can see from the chart that there exists a sharp transition in GAP score when changing the method from Logistic Regression to LSTM. It is not surprising because LR model only make use of video-level features. To be specific, it totally abandon the rich and insightful information hiding between frames. Thus the result can not be very satisfying. It is only 0.70727.
\subsection{From LSTM to Bi-LSTM}
Bidirectional LSTM overcomes the disadvantages that a single-direction network meets, obviously that the model benefits from its subtle architecture. 

Traditional LSTM is able to capture the current frame's dependencies on previous frames, which implies that the current frame is relevant to frames ahead of it. And our team intuitively think that the current frame is also largely dependent on the following frames. Therefore, the bidirectional LSTM can work better. Bidirectional LSTM networks can slightly improve the performance of classification and recognition.In the experiment, it generally performs best. 
\subsection{LSTM vs GRU}
Bi-GRU model with concat modalities showed significantly improvement in GAP score. They have similar structures, which includes many gates and hidden states, LSTM units also have cells. We presume that it is because GRU has simpler structure and less parameters, which is easier to converge. The GAP score of stacked Bi-GRU is 0.04 more than stacked Bi-LSTM. 
\subsection{Three modality fusion methods}
It is astonishing to see that GAP value of shared-space and projection method are all much lower than concat method. Simpler is better. Multiplying weight matrix for each feature vector may destruct the internal meaning of different modalities. However, the training loss is 0.79 and is just slightly lower than that of concat method. Obviously, the models suffer from overfitting. That may because the weight matrix limit the representation power of the models.
\subsection{Attention}
From the results, we can see that attention on the contrary decreases the GAP score. It seems to be a violation of the principle that complex model improves the representation power. We think that the complex model make it difficult to train, and the gap is because of training algorithm, with careful parameter choose, the model with attention mechanism would outperform the initial model.
\subsection{Ensemble method}
Because of the time limit, we only try to ensemble two models, LR and LSTM. It takes the third place among our experiments. Simple models can make effective model and beat end-to-end models, which is impressive and lead us to explore more successful ensemble algorithms.

To understand that why the performance of an ensemble-version model is usually much better than that of a single learner, we refer to Dietterich \cite{ditterrich1997machine} and think about his three reasons by regarding the nature of machine learning as searching for the most accurate hypothesis in a hypothesis space. Firstly, the training data might not provide sufficient information for choosing a single best learner. Therefore, combining these learners is a good idea. Secondly, the search processes of the learning algorithms might be imperfect. For example, even a unique best hypothesis really exists, it might be difficult to achieve because it is easy to get sub-optima. Thus, ensembles can make up for the imperfect search processes. Finally, the hypothesis space may not contain the true target hypothesis, while ensembles can give some good approximation. These are only some intuition rather than rigorous theoretical demonstrations. 
\subsection{Summary}
It is easy to see that GAP value is improving when we try to grasp more information from the video. The more information we use, the better performance we reach. Models of probable complexity are the ones that can make the best use of the information. All in all, it is the quality of data that matters.
\section{Conclusion}
In this paper, we first introduce the motivation of our research, which is using machine learning algorithms to handle large-scale videos and categorize them efficiently. After reviewing some related works, we implement several common methods such as Logistic Regression and Long Short Term Memory method. We improved the LSTM method by introducing Bi-LSTM and Bi-GRU structure based on the intuition that the current frame is also largely dependent on the following frames. Therefore, we use a recurrent network with bidirectional RNN cells to implement large-scale video classification problem with frame-level features(both visual and audio)and test the performance according to a measure called GAP(Global Average Precision). We also compare different multimodal fusion techniques, as well as attention mechanism and ensemble methods. Our model has the ability to be compatible with large-scale data and obtain a great result. Our best rank is 90 over 656 participating teams in the competition and the score we gain is \emph{0.80583}. After the competition, by using some other method, we finally achieve GAP score as high as 0.84022. 
\subsection{Future Work}
There are so many things we can do in the future. After communicating with other researchers, we harvest lots of feasible methods to improve the performance. For example, we can sample frames randomly to train the model instead of using the whole dataset. Sampling can narrow the redundancy between sequential frames and reduce the training time efficiently, while because of the randomness and the existing redundancy property of videos, the main information did not lose much. To get a better rank, more ingenious ensemble learning is a promising way which doesn't need much special techniques. We will also try the latest unit called SRU since GRU is much better than LSTM units, which leads to our interest in research of unit structure. Fancy ideas such as replacing Recurrent Neural Networks with Convolutional Neural Networks or Residual Networks or Transformer are also worth trying.
\newpage
\bibliographystyle{aaai}
\bibliography{aaai_bluecat}  

\end{document}